\documentclass[10pt,journal,compsoc]{IEEEtran}

\usepackage{graphicx}
\usepackage{epsf}
\usepackage{bm}
\usepackage{verbatim}
\usepackage{amsmath}
\usepackage{amssymb}
\usepackage{multirow}
\usepackage{url}
\usepackage{algorithm}
\usepackage{algorithmicx}
\usepackage{booktabs}
\usepackage{comment}

\usepackage{enumitem}
\setlist{nolistsep}
\usepackage{color}
\usepackage{subcaption}
\graphicspath{{fig/}}

\usepackage[T1]{fontenc}
\usepackage{subfig}

\usepackage{times}
\usepackage{helvet}
\usepackage{courier}
\usepackage{caption}
\usepackage[margin=0.5in]{geometry}

\renewcommand{\vec}[1]{\pmb{\bm{{#1}}}}

\newcommand{\betax}{\pmb{\bm{\beta}}}

\usepackage[noend]{algpseudocode}

\ifCLASSOPTIONcompsoc
  \usepackage[nocompress]{cite}
\else
  \usepackage{cite}
\fi

\ifCLASSINFOpdf
 \else
  \fi

\begin{document}
\title{Transfer String Kernel for Cross-Context DNA-Protein Binding Prediction}

\author{Ritambhara~Singh,
	Jack~Lanchantin,
        Gabriel~Robins,
        and Yanjun~Qi
\IEEEcompsocitemizethanks{\IEEEcompsocthanksitem All authors are with Department
of Computer Science, University of Virginia, Charlottesville
VA, 22903.\protect\\
E-mail: yanjun@virginia.edu}}

\IEEEtitleabstractindextext{

\begin{abstract} 
Through sequence-based classification, this paper tries to accurately predict the DNA binding sites of transcription factors (TFs) in an unannotated cellular context. 
Related methods in the literature fail to perform such predictions accurately, since they do not consider sample distribution shift of sequence segments from an annotated (source) context to an unannotated (target) context.
We, therefore, propose a method called ``Transfer String Kernel'' (TSK) 
that achieves improved prediction of transcription factor binding site (TFBS) using knowledge transfer via cross-context sample adaptation.  
TSK maps sequence segments to a high-dimensional feature space using a discriminative mismatch string kernel framework. In this high-dimensional space, labeled examples of the source context are re-weighted so that the revised sample distribution matches the target context more closely.  
We have experimentally verified TSK for TFBS identifications on fourteen different TFs under a cross-organism setting. We find that TSK consistently outperforms the state-of-the-art TFBS tools, especially when working with TFs whose binding sequences are not conserved across contexts. We also demonstrate the generalizability of TSK by showing its cutting-edge performance on a different set of cross-context tasks for the MHC peptide binding predictions.
\end{abstract}

\begin{IEEEkeywords}
Machine Learning, Bioinformatics, Support Vector Machines, Domain Adaptation, String Classification, String Kernel 
\end{IEEEkeywords}

This work will be originally published in IEEE/ACM Transactions on Computational Biology and Bioinformatics. Code available at github.com/QData/TransferStringKernel. }

\maketitle

\IEEEdisplaynontitleabstractindextext

\IEEEpeerreviewmaketitle

\IEEEraisesectionheading{\section{Introduction}\label{sec:introduction}}

\IEEEPARstart{S}{equence} analysis plays an important role in the field of bioinformatics. 
Genomic sequences build the basis of a large body of research on understanding the biological processes in living organisms.
As an important application of sequence based mining, the task of predicting Transcription Factor Binding Sites (TFBSs) on genomes has attracted much attention over the years \cite{encode2012integrated}.
Transcription factors (TFs) are regulatory proteins that bind on functional sites of DNA to control the regulation of genes.
Each different TF binds to specific locations (or sites) on a genomic sequence to regulate cell machinery. While many TFs are highly conserved among different species \cite{vaquerizas2009census}, conservation of their binding sequence is low \cite{dermitzakis2002evolution}. These TFBSs vary across different cell types, cell stages and genomes. 
Owing to the development of chromatin immunoprecipitation and 
massively parallel DNA  sequencing (ChIP-seq) technologies \cite{park2009chip}, maps of genome-wide binding sites are currently  available for multiple TFs in a few cell types across human and mouse genomes via the ENCODE \cite{encode2012integrated} database. Because ChIP-seq experiments are slow and expensive, they have not been performed for many important cell types or organisms. Therefore, computational methods to identify 
TFBS accurately remain essential for understanding the regulatory functioning and evolution of genomes. Such predictions are especially important for unannotated cellular contexts (e.g., cell types of rare diseases or rare organisms). 

String kernel techniques under the support vector machine (SVM) classification framework have been successfully used for detecting patterns of DNA or protein sequences before \cite{leslie2004fast}. Through local substring ($k$-mer) comparisons that incorporate mismatches, 
this category of models extracts mismatch features and trains to classify sequence segments from a set of previously labeled sequences. Then, the learned models are used to classify a new set of sequences.  Recently, \cite{arvey2012sequence} extended this discriminative SK+SVM classification setup for predicting TFBSs 
in a cell-type specific manner. Their results \cite{arvey2012sequence} have suggested that traditional motif-driven approaches (details in Section \ref{subsec:related}) for TFBS predictions are not always sufficient for accurately accounting for cell-type specific binding profiles. Despite the exciting aforementioned framework \cite{arvey2012sequence},   
one shortcoming remains significant: this method assumes that the source and target samples are drawn from the same probability distribution. This makes string kernel non-applicable to most cases of TFBS predictions in unannotated cellular contexts.

\begin{figure*}[t]
 \centerline{\includegraphics[width=\textwidth]{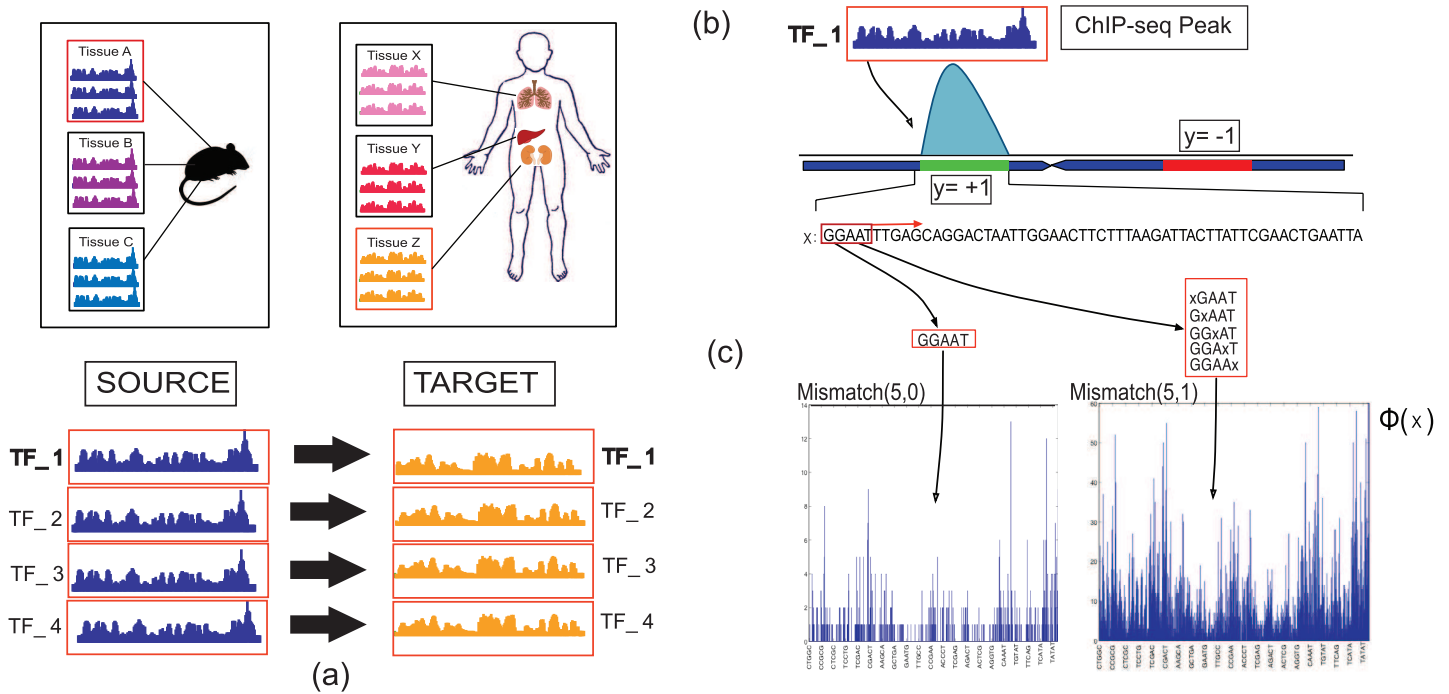}}
 \caption{Overview of our experimental setup for cross-context TFBS prediction tasks. (a) Source (mouse) and target (human) ChIP-seq maps for cross-context TFBS classifications (a family of tasks). Four example cases are shown. (b) An example of sequence based TFBS classification shown for TF\_1. ``y'' labels are obtained from ChIP-seq signals. ``x'' are corresponding sequence segments. Thus, $y \in \{+1,-1\}$, where $y=+1$ indicates binding of TF and $y=-1$ means TF does not bind to the sequence. (c) The baseline $(k,m)$-mismatch string kernel.  Here, $k=5$ and $m \in \{0,1\}$.
\label{fig:overall}}
\end{figure*}

We take into consideration that if a genome-wide ChIP-seq experiment has been performed for a specific TF in a certain context, it is normally unnecessary to computationally predict TFBSs for that specific [TF, context] pair again. Therefore, this paper focuses on computationally predicting TFBSs for cellular contexts that have not yet been annotated. We generate the ``gold standard'' reference labels, used for training, from an existing ChIP-seq TFBS map of an annotated context that is most related to the target [TF, context] configuration of interest. 
Since each TF's binding sequence patterns vary with the context factor,
it is likely that data samples in the annotated (source) context are distributed differently from samples in the unannotated (target) context. 
This also connects to observations in the literature that the TFBS sequences have low conservation across species even though many TF proteins are conserved across large evolutionary distances \cite{chen2007evolution}. When implementing cross context prediction, it is desirable to design algorithms that remain effective under such distribution shifts \cite{yu2012analysis}. 
While the problem may be unsolvable if the source and target distributions share nothing in common, recent machine learning studies 
 \cite{gretton2009covariate} have provided effective adaptation for closely related distributions. 
Biologically, this also makes sense since closely related cell contexts should have similar gene regulation processes and similar TFBS sequence patterns  \cite{arvey2012sequence}. 

In this paper, we propose an approach called ``Transfer String Kernel (TSK)'' that learns to transfer knowledge from an annotated to an unannotated cellular context to achieve better sequence-based TFBS predictions. Essentially, TSK re-weighs the source data so that its distribution more closely matches that of the target data \cite{gretton2009covariate}. It assigns higher weights to those observations in the source context that are most similar to those in the target context, and lower weights to those that rarely occur in that context. This connects to a more general machine-learning topic called {\em domain adaptation} \cite{schweikert2009empirical} (discussed in Section \ref{subsec:related}).  

In summary, this paper includes the following contributions:
\begin{itemize}
\item ``Transfer String Kernel (TSK)'' implements domain transfer of string kernel via Kernel Mean Matching,
on tasks for which target domain is unannotated;
\item Experimentally, we apply TSK on 14 different TFBS prediction tasks from mouse to human genomes. 
To the best of authors' knowledge, this is the first attempt of studying cross-context TFBS predictions across different genomes. This setting
resembles the translational setup that many biologists are working on; 
\item We demonstrate that TSK consistently outperforms state-of-the-art TFBS prediction tools, especially for those TFs whose binding properties do not conserve across contexts;
\item Our experiments show that TSK exhibits robust performance despite the label imbalance issue that is common in TFBS prediction tasks;
\item Finally, we show that TSK generalizes well to other  cross-context sequence based prediction tasks.
\end{itemize}

TSK is a general cross-context sequence modeling
approach and not tied to the TFBS applications. We show this generality by applying it for predicting peptide binding (PB) to Major Histocompatibility Complex Molecules (MHC class I). 
MHC is a set of cell surface proteins that bind to peptide fragments derived from pathogens and display them on cell surface for recognition by the appropriate T-cells (i.e., consequently controlling the adaptive immune response).
Knowledge of these peptide bindings helps in studying immunity and is vital for vaccine development. 
The protein sequence-based PB prediction tasks are similar to TFBS tasks because: a) MHC PB bindings vary across organisms and b) obtaining such experimental data is a complex, expensive and time consuming task, therefore leading to
lack of labeled samples in target (human) domain. However, it differs from TFBS prediction tasks in its larger dictionary size (due to protein sequences) as well as in the sample properties (e.g., shorter sequence length, varying ratios of positive to negative samples and size) of the datasets.

\section{Methods: Transfer String Kernel (TSK) }
\begin{table}[t]
\def\arraystretch{1.3}
\caption{Notations we use and their descriptions.}
\label{Tab:01}
\begin{tabular}{|p{1.4cm}|p{5.5cm}|} \hline
Notations & Descriptions\\ \hline
$x$ & Input sample \\ \hline
$y$ & Output label \\ \hline
$sd$ & Source domain\\ \hline
$td$ & Target domain\\ \hline
$n_{sd}$ & Number of source domain samples\\ \hline
$n_{td}$ & Number of target domain samples\\ \hline
$w$ & Decision hyperplane of the weight vector\\ \hline
$b$ & Intercept term of the hyperplane \\ \hline
$\alpha_{i}$ & Slack variable (Lagrange multiplier associated with each constraint $y_{i}=(w^{T}x_{i}+b)) \geq 1$\\ \hline
$K$ & Kernel matrix among source samples\\ \hline
$\kappa_{i}$ & Weighted sum of kernel values among a source sample and all target samples \\ \hline
$\betax$ & Importance weight vector (Dimension=$n_{sd}$)\\ \hline
\end{tabular}
\end{table}

We propose a learning method, called ``Transfer String Kernel'' (TSK) that performs cross-context ``knowledge transfer'' and utilizes available TF binding labels from a source context to discriminatively predict TFBS for the same specific TF in an unannotated target context. We now discuss various components of
our approach. Table~\ref{Tab:01} provides a list of notation symbols that we have used in this section.

\subsection{Basic Tasks: A family of sequence classification tasks}
The sequence-based TFBS prediction problem can be casted as 
a string classification task.  
For a certain TF of interest, on a particular genome under a specific cellular context, the task aims to classify a DNA sequence segment as a potential TFBS or a non-binding background site (see a sample sequence-label pair $(x,y)$ in Figure~\ref{fig:overall}(b)).
When considering multiple TFs and various cellular contexts (Figure~\ref{fig:overall}(a)), 
this constitutes a family of sequence classification problems. For example in Figure~\ref{fig:overall}(b), we train a string classifier using labeled segments (obtained from ChIP-seq) of TF\_1 from the source context. The classifier is then used to predict potential TFBS sites of the same TF (TF\_1) in the target context. This means that if we work on $N$ different TFs, we are
handling a family of $N$ different classification tasks. In case of Figure \ref{fig:overall}(a), there are four different TFBS tasks for 4 TFs.

For each task, when performing sequence-based TFBS predictions in an unannotated cellular context, we face a data problem called {\em distribution shift}. We first define necessary notations to explain this problem. We have two spaces,  $\mathcal{X}$ represents all possible sequence patterns and $\mathcal{Y}$ represents their respective labels. The set of source samples:

\begin{equation}
Z_{sd}=\{(x_1^{sd}, y_1^{sd}), ..., (x_{n_{sd}}^{sd}, y_{n_{sd}}^{sd})\} \subseteq \mathcal{X} \times \mathcal{Y},
\end{equation}
follow a probability distribution $P_{sd}(x,y)$ \footnote{This probability varies with different $TF_i$} in the annotated source context. Target samples (for the same $TF_i$) form the following set: 
\begin{equation}
Z_{td}=\{(x_1^{td}, y_1^{td}), ..., (x_{n_{td}}^{td}, y_{n_{td}}^{td})\} \subseteq \mathcal{X} \times \mathcal{Y},
\end{equation}
drawn from another distribution $P_{td}(x,y)$ in the unannotated target context. In the training phase, we are assuming a transductive setting, in which we can access a set of $x_{i}^{td}$. True label, $y_{i}^{td}$, is not available for training and is used in testing for evaluation. 

\subsection{Basic Model: Mismatch String Kernels with SVM}
\label{Sec:Spectrum}
 The key idea of string kernels is to apply a function $\phi(\cdot)$, which
maps sequences of arbitrary length into a vectorial feature space of fixed length.
In this space, a standard classifier such as a support vector machine (SVM) ~\cite{VAPNIKLEARNING} can then be applied.
Kernel-version SVMs 
calculate the decision function for an input sample $x$ as,
\begin{equation} \label{eq:fsvm}
f(x) = \sum_{i \in sd}  \alpha_i K(x_i^{sd}, x) +b.
\end{equation}
String kernels \cite{leslie2004fast}\cite{kuksa2008scalable}, implicitly compute an inner product in the mapped feature space $\phi(x)$ as:
\begin{equation}
K(x,x') = \langle \phi(x), \phi(x') \rangle,
             \label{eq:kernel}
\end{equation}
where $x = (s_1, \dots, s_{|x|})$,
$x,x' \in \mathcal{S}$, 
$|x|$ denotes the length of the sequence $x$, $\mathcal{S}$ represents 
the set of all sequences composed 
of all possible dictionary items (for example, in case of DNA sequences $s_i \in \{ A,C,T,G \} $), and
$\phi: \mathcal{S} \to R^m$ defines the mapping from a sequence $x \in \mathcal{S}$ to a $m$-dimensional feature vector.

The feature representation $\phi(\cdot)$ plays a key role in the 
effectiveness of sequence analysis since biological sequences cannot be
readily described as feature vectors.
One classic representation is
to treat DNA sequence segments as an unordered set of nucleotide $k$-mers 
(combinations of $k$ adjacent nucleotide residues). A feature
vector indexed by all $k$-mers records the number of occurrences of
each $k$-mer in the current sequence.   
The string kernel using this representation is called {\em spectrum
  kernel}~\cite{leslie2004fast}, where the spectrum representation 
counts the occurrences of each nucleotide $k$-mer in
a DNA sequence segment.
Kernel scores between sequences are then
computed by taking an inner product between corresponding ``$k$-mer - indexed'' feature vectors:
\begin{equation}
K(x,x') = \sum_{\gamma \in \Gamma_k}
             c_x(\gamma) \cdot c_{x'}(\gamma),
\end{equation}
where $\gamma$ represents a $k$-mer, $\Gamma_k$ is the set of all
possible $k$-mers, and $c_x(\gamma)$ ~is the number of occurrences (with
normalization) of $k$-mer $\gamma$ in sequence $x$.

However, exact kernel matching in biological sequences is ineffective due to naturally occurring letter (e.g., nucleotide) substitutions, insertions, or deletions. Therefore, inexact comparison is critical for effective matching between sequence segments. In string kernels, this is typically achieved by using different families of mismatches \cite{kuksa2008scalable}. 
We use the concept of $(k,m)$-mismatch string kernels \cite{kuksa2008scalable} (illustrated in Figure~\ref{fig:overall}(c)), which considers $k$-mer counts with $m$ inexact matching of
$k$-mers (e.g. $k$=5, $m \in \{0,1\}$). For this kernel, 

\begin{align}
K(x,x') = \sum_{\gamma \in \Gamma_k} c_x^{k,m}(\gamma) \cdot c_{x'}^{k,m}(\gamma),\\
c_x^{k,m} = \sum_{g \in S(\gamma,m,k)}c_x(g),     
\end{align}

$S(\gamma,m,k)$ denotes the set of contiguous substrings
of length $k$ that differ from $\gamma$ in at most $m$ positions (see Figure~\ref{fig:overall}(c) with an example case for $(k=5, m \in \{0,1\})$).

Through mismatches, $(k,m)$-mismatch kernel allows flexibility when matching $k$-mers between sequences. This flexibility partly reflects the true biological nature of DNA or proteins sequences since they are prone to mutations like deletions, insertions, substitutions etc. As expected, it performs better than spectrum kernel (as seen in Section 4.1) therefore; it is a natural choice for our TSK approach.

In practice, string kernel implementations typically require efficient 
computation without explicitly constructing potentially high-dimensional feature vectors $\phi(\cdot)$. This is because the explicit feature
vector mapping $\phi(\cdot)$ becomes problematic (especially in the case of inexact matching) for even small value of $k$ (due to the dimensionality of
$\phi(\cdot)$ being exponential in $k$). 
Researchers have proposed various strategies~\cite{leslie2004fast,kuksa2008scalable}
to address these computational difficulties. We adopt
a statistical strategy from~\cite{kuksa2008scalable} that provides linear-time string kernel computations and scales well with dictionary size and input length.  

\subsection{Proposed Model: Transfer String Kernel }
String kernel assumes training and testing samples are drawn from the same probability distribution. To consider the variation between source and target samples in our application, {\em domain adaptation} \cite{pan2010survey,schweikert2009empirical} serves as a natural candidate to tackle this computational challenge. In machine learning, {\em domain adaptation} aims to use data or a model of a well-analyzed source domain to obtain or refine a model for a less analyzed target domain. The  
specific ``domain transfer'' setting being focused in this paper assumes that 
 historical labels only exist in the source context and 
are not available in the target context (reasons explained in Section 1).  Accordingly, we propose ``Transfer String Kernel (TSK)'' approach to achieve better cross-context TFBS predictions by transferring knowledge from an annotated context to an unannotated context.

TSK revises the $(k,m)$-mismatch string kernel framework using a ``Kernel Mean Matching'' (KMM) strategy \cite{gretton2009covariate} in order to
perform knowledge transfer. 
To our knowledge, TSK has not been proposed in the literature for cross-context TFBS prediction tasks.
Specifically, TSK adapts string kernel under
the ``covariate shift'' assumption \cite{gretton2009covariate}. It assumes that
 the conditional probability distribution of the output variable,
given the input variable, remains fixed in both the source and
target set. In other words, the data shift happens for the marginal
probability distribution of feature variables (from $P_{sd}(x)$ to $P_{td}(x)$) and not for the conditional
distribution $P(y|x)$.  

The key to correcting this type of sampling bias is to estimate the ``importance weight'' for each source sample \cite{yu2012analysis}:  
\begin{align}
\beta (x,y) := \frac{P_{td}(x,y)}{P_{sd}(x,y)} = \frac{P(y|x)P_{td}(x)}{P(y|x)P_{sd} (x)} = \frac{P_{td}(x)}{P_{sd} (x)}.
\label{eq:ratio}
\end{align}

The KMM estimator accounts for the difference between $P_{td}(x,y)$ and $P_{sd}(x,y)$ by re-weighting the source points so that the means of the source and target points in a Reproducing Kernel Hilbert Space (RKHS) are close.  This reweighing process is therefore called ``Kernel Mean Matching'' (KMM). When the RKHS is universal, the population solution of weight vector $\hat{\betax}$ (in the following Equation~\ref{eq:beta1}) to the KMM optimization (of matching means) is exactly the vector form of ratio $\frac{P_{td} (x)}{P_{sd} (x)}$ (in Equation~\ref{eq:ratio}) including all $x$ from the source domain. Let $\beta_i$ represent the ``importance weight'' of a source instance $x_i$. More specifically, KMM uses a ``maximum mean discrepancy'' measure to minimize the difference between the empirical mean of the source and the empirical mean of the target distribution.  Formally, KMM attempts to match the mean elements in a feature space induced by a kernel function $K(\cdot,\cdot)$:  

\begin{align}
\label{eq:beta1}
\hat{\betax} & = argmin_{\betax} \hat{L}(\betax),
\end{align}

where, 

\begin{align}
\hat{L}(\betax)  & = \Arrowvert \frac{1}{n_{sd}} \sum_{i=1}^{n_{sd}}{\beta}_i * \phi(x^{sd}_i) - \frac{1}{n_{td}} \sum_{i=1}^{n_{td}} \phi(x^{td}_i)\Arrowvert ^2 \\
& = \frac{1}{n^2_{sd}} {\betax}^T K {\betax} - \frac{2}{n^2_{sd}} \kappa^T {\vec{\beta}} + \text{constant}, \\
& s.t. \quad \beta_i \in [0, B] \qquad \text{and} \qquad \lvert {\sum_{i=1}^{n_{sd}} {\beta_i} - n_{sd}}\rvert  \leq n_{sd} \epsilon \nonumber.
\end{align}

Here, $K$ is the kernel matrix among all source examples, and $\kappa_{i}$ is the weighted sum of kernel values among a source example and all target examples:
\begin{align}
K_{ij}:= & K(x^{sd}_i, x^{sd}_j),  \\  
\kappa_i := & \frac{n_{sd}}{n_{td}}\sum^{n_{td}}_{j=1} K(x^{sd}_i, x^{td}_j).
\end{align}
Large values of $\kappa_i$ correspond to more important observations $x^{sd}_i$ and are more likely to lead to larger $\beta_i$.  
Equation~\ref{eq:beta1} involves a quadratic program which can be efficiently solved using the interior point methods or any other subsequent procedures such as projected gradient optimization method \cite{gretton2009covariate}. 

The derived importance weights $\hat{\betax}$ are then used to re-weigh source samples in a revised support vector machine training algorithm, i.e., an instance weighted SVM that aims to optimize: 
\begin{align}
\sum^{n_{sd}}_{i=1} & \alpha_i - 0.5 \sum_{i,j \in sd} \alpha_i \alpha_j y_i y_j K(x^{sd}_i, x^{sd}_j), \\
s.t. & \hspace{2mm} \sum^{n_{sd}}_{i=1}  \alpha_i y_i = 0  \text{,} \hspace{3mm}   
\text{and} \hspace{3mm} \hat{\beta}_i C \geq
\alpha_i \geq 0.
\end{align}
Here $\alpha_i$ are the slack variables in the SVM, the same as $\alpha_i$ in Eq~\ref{eq:fsvm}.  

In summary, we apply KMM in conjunction with the $(k,m)$-mismatch string kernel using the algorithm summarized in Algorithm~\ref{alg:tskm}, thus enabling us to perform knowledge transfer across domains when classifying strings.

\begin{algorithm}[t]
\caption{Algorithm for Transfer String Kernel through matching means in RKHS}\label{alg:tskm}
\begin{algorithmic}[1]
\Procedure{TSK}{}
\State Implement $(k,m)$-mismatch string kernel to calculate the kernel matrix $K$ and vector $\kappa$; 
\State Use Kernel Mean Matching estimator on $K$ and $\kappa$ to obtain the importance weight $\hat{\beta}_i$ for each source sample $i$;  
\State Use $\hat{\beta}_i$ to perform instance re-weighted SVM training step; 
\State Use the trained SVM model to perform sequence classification of target samples. 
\EndProcedure
\end{algorithmic}
\end{algorithm}

\subsection{Connecting to Previous Studies}
\label{subsec:related}

\subsubsection{Sequence-motif based TFBS prediction tools} 
Transcription factors influence gene expression by binding to specific DNA sequence in a genomic region. Therefore accurate models for identifying and describing the binding sites of TFs are essential in understanding cells. Previous techniques include many sequence-motif based computational approaches that typically use position-based sequence information for predicting TFBS. Relying on a set of known transcription factor binding sites (TFBSs) for a given TF, the binding preference is generally represented in the form of a position weight matrix (PWM) \cite{stormo2013modeling,mathelier2013jaspar} (also called position-specific scoring matrix) derived from a position frequency matrix (PFM). A PFM is essentially an occurrence table, summarizing the number of each nucleotide being observed at each position of a set of aligned TFBSs.  
\cite{arvey2012sequence} have suggested that traditional motif-driven approaches are not always sufficient to accurately account for cell-type specific binding profiles.  While motif-based PWMs are compact and interpretable, they can under-fit ChIP-seq data by failing to capture subtle but detectable and important sequence signals, such as direct DNA-binding preferences of certain TFs, cofactor binding sequences, accessibility signals, or other discriminative sequence features. 
Gene regulatory programs are primarily regulated through cell-context specific binding of TFs. Therefore, it becomes more clear that TFBS analyses should consider distinct cell contexts (see Figure~\ref{fig:overall}(a)).  
To the best of authors' knowledge, this paper is the first attempt of direct testing of cross-context prediction performance of such methods. 

\subsubsection{String Classification}

Our formulation of TFBS prediction belongs to a general category of "string classification".
Various methods have previously been
proposed to solve string classification, 
including generative (e.g., Hidden Markov Models-HMMs) and discriminative approaches. 
Among the discriminative approaches, string kernel methods
provide some of the most accurate results, such as for remote protein fold and homology
detections~\cite{leslie2004fast,kuksa2008scalable}. 
Aside from spectrum kernel~\cite{leslie2004fast} and $(k,m)$-mismatch string kernel \cite{kuksa2008scalable} introduced in the previous section, a few other 
notable string kernels include (but are not limited to): (1) The gapped kernel calculates dot-product of (non-contiguous) $k$-mer counts with gaps allowed between elements. (2) More generally, the substring kernel~\cite{vishwanathan2004fast} 
 measures similarity between sequences based on common
co-occurrence of exact sub-patterns (e.g., substrings). (3) 
The profile kernel ~\cite{kuang2005profile} uses the notion of
similarity based on a probabilistic model (e.g. profile). 
(4) Under the semi-supervised setting, the so-called ``sequence neighborhood'' kernel or ``cluster'' kernel ~\cite{chapelle2002cluster} proposes a semi-supervised extension of string kernel, which replaces every sequence with a set of ``similar'' (neighboring)
sequences and a new representation is obtained by averaging representations of
these neighboring sequences found in the unlabeled data using a certain sequence similarity measure.

Sequence labeling or sequence tagging is another related category of formulations for sequence modeling.
Many successful works in the
field of natural language processing \cite{collobert2008unified} belong to this category in which each position of input sequences can
be annotated with tags indicating parts of
speech, named entities, semantic roles, etc. Popular bioinformatics tasks predicting proteins' local functional properties \cite{qi2012unified} have been modeled as a labeling or tagging of amino acids on proteins.  Many important functional properties are computationally predicted in
this fashion, such as secondary structure, solvent accessibility, 
transmembrane topology or the locations of coiled-coil
regions. Multiple classic machine learning methods have been benchmark tools for sequence tagging, like conditional random field \cite{lafferty2001conditional}. We omit a full survey of this topic due to its vast body of previous literature and loose connection to our TFBS formulations. 

\subsubsection{Domain Transfer for Genome Sequence Mining}
TSK relates to a more general machine-learning topic ``covariate shift". 
Traditional supervised learning assumes source
and target samples are usually drawn from the same probability distribution. 
This assumption can be easily violated in practice, for instance, due to sampling bias or nonstationarity of the sampling environment (i.e. the case of TSK) \cite{yu2012analysis}. Algorithms that remain effective under such distribution shifts are highly preferred. This problem has been 
investigated in both statistics and machine learning under various  assumptions \cite{yu2012analysis}. The covariate shift assumption assumes that the conditional probability distribution of the output variable $y$, given the input variable, $x$ remains fixed in both the source and target. 
Under this setting, the basic motivation to correct the sampling bias 
 is to re-weigh the source data so that its distribution matches 
 more closely to that of the target data. A number of previous methods have been proposed to estimate the ``importance weight'' $\betax$ from finite samples, including kernel mean matching (KMM), logistic regression, KL importance estimation and many others  \cite{yu2012analysis}. 
 
More generally, TSK belongs to topics of {\em domain adaptation} and {\em transfer learning}\cite{pan2010survey} \cite{schweikert2009empirical}, where one aims to use data or models of a well-analyzed source domain to obtain or refine models for a less analyzed target domain. 
This helps when datasets from the two domains belong to different feature spaces or follow different data distributions. For such scenarios, knowledge transfer will potentially improve the performance of learning by avoiding or reducing the expensive data-labeling efforts \cite{pan2010survey}. 
A good classifier can only be trained when a sufficient
amount of annotated training data is available. However, labeled training data in biomedical fields is scarce.
Obtaining such labeled datasets can be costly and time-consuming. This results in the data sparsity problem that is a major bottleneck for
applying machine learning in biomedical domain. 
Computational methods with the ability of ``knowledge transfer'' 
become highly crucial for such applications because biomedical data is intrinsically complex and heterogeneous at almost every level (e.g., different
species, different tissue types, etc). Transfer learning can help in reusing knowledge
from annotated datasets to new domains \cite{xu2011survey}, reducing the knowledge gap of labeled data due to heterogenous variations.
For example, one previous study \cite{schweikert2009empirical} explored ``domain adaptation'' strategies 
for a task of sequence-based prediction for acceptor splice sites. 
It assumed that historical labels exist for both source and target contexts, which is different from our setting. We aim to predict TFBS for a cellular context that has not yet been annotated (i.e. no label exists in the target domain). 
Our setting is more practical for TFBS prediction tasks, because if a ChIP-seq experiment has been performed for a specific TF in a certain context, it is normally unnecessary to computationally predict TFBS for that specific domain again. 
Experimental results in Section 4 successfully present a strong case of mining bio-medical data where the transfer of knowledge across domains is critical and fruitful.

\begin{table*}[t]
\centering
\captionsetup{justification=centering} 
\caption{Sample sizes and hyperparameters in TFBS prediction experiments.}
\label{Tab:02}
{\begin{tabular}{ | c | c | c | c | c | c | c | c |}
\hline
	\multicolumn{3}{ | c }{\textbf{Source Samples : Mouse (train)} }&  
	\multicolumn{3}{ | c }{\textbf{Target Samples : Human (validation and test)}} &  
	\multicolumn{2}{ | c | }{\textbf{Kernel Parameters}}  \\ \hline
	Pos sites & Neg sites & Total (n) & Pos sites & Neg sites & Ratios (r) & k & m \\ \hline
	500 & 500 & 1000 & 500 & 500,1000,1500 & 1:1,1:2,1:3 & 8, 10 ,12 & 1,2,3 \\ \cline{1-5} \hline
\end{tabular}}
\end{table*}

\begin{table*}[t]
\centering
\captionsetup{justification=centering} 
\caption{Sample sizes and hyperparameters in PB prediction experiments.}
\label{Tab:03}
{\begin{tabular}{ | c | c | c | c | c | c | c | c | c | c | c | c |}
\hline
	\multicolumn{3}{ | c }{\textbf{Source Samples: Mouse (train)} }&  
	\multicolumn{4}{ | c }{\textbf{Target Samples: Human (validation)}} & 
	\multicolumn{3}{ | c }{\textbf{Target Samples: Human (test)}} &  
	\multicolumn{2}{ | c | }{\textbf{Kernel Parameters}}  \\ \hline
	Pos sites & Neg sites & Total (n) & Target & Pos sites & Neg sites & Total (n) & Pos sites & Neg sites & Total (n) & k & m \\ \hline
	\multirow {5}{*}{586} & \multirow {5}{*}{2598} & \multirow {5}{*}{3184} & HLA-A0201 & 1370 & 11752 & 13122 & 82 & 1169 & 1251 & \multirow {5}{*}{5,6,7} & \multirow {5}{*}{1,2,3} \\ \cline{4-10}
	& & & HLA-B0702 & 170 & 3627 & 3797 & 135 & 1592 & 1727 & &\\ \cline{4-10} 
        & & & HLA-B4403 & 204 & 538 & 742 & 91 & 629 & 720 & &\\ \cline{4-10} 
        & & & HLA-B5301 & 121 & 806 & 927 & 109 & 376 & 485 & &\\ \cline{4-10} 
	& & & HLA-B5701 & 237& 1882 & 2119 & 286 & 550 & 836 & &\\ \cline{4-10} \hline
	\end{tabular}}
\end{table*}

\begin{table*}[t]
\centering
\caption{Comparison using average test AUC scores of 14 TFs for spectrum (Spec.) kernel and $(k,m)$-mismatch (Mis.) kernel. $(k,m)$-mismatch kernel consistently outperforms spectrum kernel, making it the natural choice for TSK approach.}
\label{Tab:04a}
\begin{tabular}{| c | c | c | c | c | c | c | }
	\cline{1-6}
	\multicolumn{2}{| c |}{\textbf{Ratio=1:1}} &
	\multicolumn{2}{ c | }{\textbf{Ratio=1:2}} &
	\multicolumn{2}{ c |}{\textbf{Ratio=1:3}} \\ 
	\cline{1-6}
	Mis. Kernel & Spec. Kernel & Mis. Kernel & Spec. Kernel & Mis. Kernel & Spec. Kernel \\ 
	\cline{1-6}
	 \textbf{0.7786} & 0.7694 & \textbf{0.7804} & 0.7731 & \textbf{0.7805} & 0.7734 \\	
	\cline{1-6}
\end{tabular}
\end{table*}

\subsubsection{Covariate Shift \& Domain Adaptation}

Domain adaptation has been widely studied in machine learning literature to train an effective classifier for a target domain for which labeled data is rare,  or even unavailable (this paper's case). By exploiting labeled training samples from a different but related source domain, a variety of previous studies have demonstrated its effectiveness on multiple applications including sentiment analysis \cite{shi2012information}, object classification \cite{gopalan2011domain}, activity recognition \cite{hachiya2012importance},  text categorization \cite{long2015domain} and Brain-computer interface (BCI) tasks \cite{li2010application}. Roughly speaking, previous learning methods for domain adaptation can be summarized into two main types:

\textbf{Covariate Shift:} First, many existing studies try to estimate weights of samples that account for the mismatch in distributions between a target and a source domain. Covariate shift assumes that the marginal distributions of the source and target instances differ while the conditional distribution of the target output remains the same across domains \cite{shimodaira2000improving}.  Methods for domain adaptation under covariate shift assumptions mostly try to estimate the weights of source instances so that the weighted source distributions are most similar to the target distribution. Then, an instance-weighted classifier is trained for the target domain.
For instance, the authors of \cite{dudik2005correcting}
propose density estimators that incorporate sample selection bias to adapt two distributions. The Kernel Mean Matching (KMM) method \cite{huang2007correcting, gretton2009covariate} then learns the weights by matching the distributions in a RKHS, with a recent extension \cite{zhang2013covariate} developing surrogate-kernel based kernel matching. Later, \cite{sugiyama2007covariate,tsuboi2009direct} propose to minimize the Kullback-Leibler divergence between the target distribution and the weighted source distribution. 
Several recent papers \cite{kanamori2009least,hachiya2012importance,yamada2013relative} aim to 
estimate the optimal weights by solving least-square based formulations. Furthermore, theoretical analysis of this type of domain adaptation has been studied by \cite{cortes2010learning}.

The aforementioned studies separate the estimation of importance weights and the training of weighted learning models in two stages. Several existing machine learning methods have been explored in this framework, such as importance-weighted logistic regression \cite{tsuboi2009direct}, importance-weighted kernel regression \cite{sugiyama2007covariate}, importance-weighted Gaussian processes \cite{yamada2012no}, and the so-called consistent distance metric learning method \cite{cao2011distance}.

Moreover, a few studies try to unify the two stages. For instance, the authors of \cite{bickel2009discriminative} propose a discriminative learning based adaptation that trains an integrated model to obtain both importance weights and the classifier at the same time.
As another example, a so-called doubly robust covariate shift correction method \cite{reddi2014doubly} first trains an non-weighted classifier model, estimates the weights of source instances, and then retrains an instance-weighted classification model with the learned weights. 

\textbf{Subspace Adaptation:} Another direction of  domain adaptation is to extract a subspace, or newer feature representations in the data space, to model invariant parts across target and source distributions. Transferring knowledge between domains is through such learned subspaces or feature representations. For example, \cite{daume2009frustratingly,shi2012information} propose novel feature representations for the domain adaptation purpose. Transfer component analysis \cite{pan2011domain}
was introduced to find low dimensional representations in RHKS where the target and source domains are similar. The authors of \cite{si2010bregman} try to learn a linear subspace that is suitable for knowledge transfer by minimizing Bregman divergence between target and source distributions in this subspace. A related study \cite{shao2012low} transforms target samples such that they are a linear combination of basis from the source domain. More recently, \cite{muandet2013domain} learns a domain-invariant data transformation to minimize differences between source and target distributions while preserving functional relations between data and labels. Furthermore, the authors of \cite{fernando2014subspace} propose to identify subspaces that align to the eigenspaces of the target and source domains.

\subsubsection{Sequence-based Prediction of peptide binding to MHC}

As mentioned in the introduction, the proposed TSK method is general to any cross-domain sequence modeling problems. We implement TSK on another sequence-based bioinformatics task for MHC peptide binding and compare TSK with state-of-art tools on benchmark datasets. 

The first machine learning competition in Immunology (2011) ~\cite{zhang2011machine} compared various computational algorithms for classifying peptide binding versus non-binding sites for multiple MHC molecules.
They used experimental labels of peptide binding from 3 classes of MHC molecules in humans as performance benchmarks.
In 2012, the second machine learning competition in Immunology \cite{mlcompimmune} increased the number of experimental datasets for both human and mouse molecules, providing training and test data for both species.
The task was formulated as classifying eluted peptide (naturally processed by MHC) as positive samples while simple binding and non-binding peptides
were classified as negative labeled samples.

For peptide-protein interaction prediction, relative positions of amino acids
and their physiochemical properties play a very important role ~\cite{giguere2013mhc}. 
The oligo kernel proposed by ~\cite{meinicke2004oligo}, takes the relative positions into consideration by 
assigning a weight to each common k-mer based on their relative position in the two peptide sequences.
For the 2012 Machine Learning Competition in Immunology for peptide binding prediction, ~\cite{giguere2013mhc}
showed that their Generic String (GS) kernel gave state-of-the-art performance when implemented on the datasets
provided by the competition from ~\cite{zhang2012toward}. According to the authors, the GS kernel includes information regarding both physiochemical 
property and relative position of amino acids during k-mer comparison. We demonstrate in Section 4 that our basic mismatch string kernel (SK) yields better performance than GS kernel on the benchmark data from this competition.

\section{Experimental Setup}
\label{sec:exp}

\subsection{TFBS Prediction Task} 
We chose a family of cross-genome TFBS prediction tasks for evaluations. 
This set of tasks transfer knowledge from mouse genome to human genome, which is the translational research setting. For a certain TF, we train a TSK model using existing ChIP-seq TFBS data of a certain cell-type of mouse genome (source) and perform TFBS predictions of the same cell-type on human genome (target). 

\subsection{Datasets} 
There exist thousands of TFs that are common between mice and humans. The ENCODE \cite{encode2012integrated} database,
however, contains only 14 ChIP-seq TF experiments that are available for both species.
Therefore, we use these 14 available ChIP-seq datasets for our cross-context predictions. The cell-type for each of the different TFs belongs to normal blood tissue, as it has the maximum number of common TF experiments across the two genomes.

\begin{table*}[t]
\centering
\caption{Comparison using test AUC scores across 14 TFBS prediction tasks. TSK outperforms SK and baselines for the majority of cases. Position Weight/Frequency Matrix based approaches, MEME and CISFINDER, also perform well for three TFs that have strong sequence conservation across genomes. The last column ``CS'' represents the conservation score of the corresponding TF. We note that SK and TSK perform the best when the conservation score is low (11 out of 14 cases).}
\label{Tab:04}
\setlength{\tabcolsep}{0.99em}
\def\arraystretch{1.3}
\begin{tabular*}{\textwidth}{ | c | c | c | c | c | c | c | c | c | c | c | c | c | c |}
	\cline{1-13}
	&\multicolumn{4}{ c |}{\textbf{Ratio=1:1}} &
	\multicolumn{4}{ c | }{\textbf{Ratio=1:2}} &
	\multicolumn{4}{ c |}{\textbf{Ratio=1:3}} \\ 
	\cline{1-14}
	TF & TSK & SK & CISF & MEME & TSK & SK & CISF & MEME & TSK & SK & CISF & MEME & \textbf{CS} \\ 
	\cline{1-14}
	Maz & \textbf{0.9167} & 0.9102 & 0.8396 & 0.8579 & \textbf{0.9106} & 0.9052 & 0.8422 & 0.8469 & \textbf{0.9091} & 0.9038 & 0.8417 & 0.8406 & 0.0580 \\ 
	\cline{1-14}
	Jund & \textbf{0.6008} & 0.5958 & 0.5475 & 0.5861 & \textbf{0.6010} & 0.5943 & 0.5506 & 0.5955 & \textbf{0.6033} & 0.5951 & 0.5549 & 0.6027 & 0.0663 \\ 
	\cline{1-14}
	Mafk & \textbf{0.8317} & 0.8313 & 0.6038 & 0.6694 & \textbf{0.8344} & 0.8365 & 0.6111 & 0.6714& 0.8416 & \textbf{0.8456} & 0.6143 & 0.6740 & 0.0753 \\ 
	\cline{1-14}
	Tbp & \textbf{0.6305} & 0.6214 & 0.5146 & 0.5236 & \textbf{0.6325} & 0.6187 & 0.5106 & 0.5343 & \textbf{0.6283} & 0.6167 & 0.5105 & 0.5316 & 0.0827 \\ 
	\cline{1-14}
	Max & \textbf{0.8726} & 0.8642 & 0.8478 & 0.2105 & \textbf{0.8856} & 0.8747 & 0.8513 & 0.2048 & \textbf{0.8877}& 0.8759 & 0.8499 & 0.1986 & 0.0844 \\ 
	\cline{1-14}
	Ctcf & \textbf{0.8314} & 0.8252 & 0.7667 & 0.8135 & \textbf{0.8311} & 0.8282 & 0.7677 & 0.8142 & \textbf{0.8301} & 0.8268 & 0.7655 & 0.8112 & 0.0873 \\ 
	\cline{1-14}
	Chd1 & \textbf{0.6731} & 0.6677 & 0.5732 & 0.3375& \textbf{0.6724} & 0.6625 & 0.5709 & 0.3455& \textbf{0.6640} & 0.6592 & 0.5746 & 0.3457 & 0.0965 \\ 
	\cline{1-14}
	Mxi1 & \textbf{0.8630} & 0.8523 & 0.7785& 0.2522 & \textbf{0.8606} & 0.8505 & 0.7780 & 0.2476& \textbf{0.8612} & 0.8494 & 0.7794 & 0.2506 & 0.1071 \\ 
	\cline{1-14}
	P300 & 0.6288 & \textbf{0.6503} & 0.5055 & 0.5795 & 0.6375 & \textbf{0.6470} & 0.50344 & 0.5689 & 0.6412 & \textbf{0.6526} & 0.5000 & 0.5781 & 0.1235 \\ 
	\cline{1-14}
	Chd2 & \textbf{0.8355} & 0.8270 & 0.6079 & 0.7420 & \textbf{0.8383} & 0.8315 & 0.6066 & 0.7405 & \textbf{0.8460} & 0.8383 & 0.6078 & 0.7430 & 0.1236 \\ 
	\cline{1-14}
	Sin3a & \textbf{0.8821} & 0.8646 & 0.8025 & 0.2118 & \textbf{0.8841} & 0.8679 & 0.8005 & 0.2048 & \textbf{0.8805} & 0.8657 & 0.7990 & 0.2036 & 0.1254 \\ 
	\cline{1-14}
	Usf2 & 0.8812 & 0.8673 & 0.8841 & \textbf{0.9658} & 0.8856 & 0.8723 & 0.8833 & \textbf{0.96346} & 0.8815 & 0.8705 & 0.8826 & \textbf{0.9608} &  \textbf{0.1300} \\ 
	\cline{1-14}
	Rad21 & 0.7408 & 0.7176 & 0.7654 & \textbf{0.8310} & 0.7456 & 0.7214 & 0.7612 & \textbf{0.8302} & 0.7381 & 0.7164 & 0.7627 & \textbf{0.8313} &  \textbf{0.1577} \\ 
	\cline{1-14}
	Smc3 & 0.8277 & 0.8052 & 0.9113 & \textbf{0.9818} & 0.8365 & 0.8158 & 0.9150 & \textbf{0.9812} & 0.8305 & 0.8114 & 0.9163 & \textbf{0.9816} &  \textbf{0.1637} \\ 
	\cline{1-14}
	\textbf{AVG} & \textbf{0.7868} & 0.7786 & 0.7106 & 0.6116 & \textbf{0.7897} & 0.7804 & 0.7109& 0.6106 & \textbf{0.7888} & 0.7805 & 0.7114 & 0.6109 \\ 
	\cline{1-13}
\cline{1-13} \end{tabular*}
\end{table*}

\subsubsection{Sequence selection method}
To select training and testing sequences from the ENCODE database, we use the ``peak-centric standard'' as described in \cite{cuellar2012epigenetic}. According to this standard, there exists a single central position in each ChIP-seq region that represents positive binding site for a certain TF, and all other genomic positions are negative sites. Consequently, for positive sequences, we use the 100 nucleotides surrounding each peak position. For negative samples, since there is a lack of experimental evidence showing where the negative ``peak'' is, we randomly select 100 basepair-length subsequences from the genomic regions where positive ChIP-seq TFBS peaks are absent \footnote{We first select ``open'' DNA segments where the TFBS binding can take place. Next, we remove all possible TFBS regions for a particular TF. }. Coordinates of TFBS peak positions are obtained from the ChIP-seq data available in the ENCODE repository \cite{encode2012integrated}.

\subsection{Setup} 
Our experiments include a few important data statistics and hyperparameters to tune:
\begin{itemize}

\item \textbf{Dictionary size (d)}: 
Our TFBS prediction uses DNA sequence segments as input data. These strings are made up of 4 characters (also known as nucleotides): A,T,C,G. The dictionary size is $d$=4.

\item \textbf{Number of source samples (n):}
For each of the 14 TFs, we use its top 500 (ranked according to ChIP-seq peak enrichment) mouse TFBS sequences. For each TF task, we randomly select 500 negative samples using the strategy in Section 3.2.1. Totally, for each TF, we have a training set containing $n$=1000 sequences (positive=500, negative=500). 

\item \textbf{Ratio of positive to negative target samples (r)}:
In a target context, we generated a validation set for hyperparameter tuning and a test set for performance evaluation. For each of the 14 TFs, we use the subsequent 500 human TFBS sequences each for both the positive validation set and positive testing set. 
However, the number of negative validation and testing samples may vary. 
For TFBS prediction on an unannotated context, it is not practical to assume that the ratio between positive and negative sites is always balanced. This is because the number of bound sites varies from
one cell-context to another cell-context as well as across different TFs. Furthermore, the number of TFBSs is generally much smaller compared to regions without the TFBSs.
Accordingly, we create target sets in which positive to negative ratio varies, simulating real-world predictions on unlabeled datasets
\footnote{Intuitively, searching TFBS sites should consider all genomic positions. In reality, researchers use epigenomic evidence to filter out non-possible binding sites.}. 
By varying negative samples in the target context and 
keeping the positive samples constant,
we obtain different positive to negative ratios in target data, i.e r=1:1, 1:2 and 1:3 respectively. A ratio of 1:3 means we expect 3 negative sites for every observed positive site in the target data. More specifically, we generated three cases of validation and test sets for each TF task, containing n=1000 (positive=500, negative=500), n=1500 (positive=500, negative=1000) and n=2000 (positive=500, negative=1500) sequences.

\item \textbf{String kernel hyperparameters ($k$,$m$)}:
For both SK and TSK, we tune two hyperparameters: the length of input k-mer, 
$k \in \{8,10,12\}$, and the number of allowed mismatches, $m \in \{1,2,3\}$. We first perform hyperparameter selection using
our validation sets and then evaluate model performance using the selected hyperparameters on the test sets.
Hyperparameters for the SVM training ($C \in \{0.1,1,10,100,1000\}$) are also selected using the validation sets.

\end{itemize}

The setup described above is novel. Our cross-context setup differs from previous TFBS prediction works and closely resembles a real biological scenario. The data statistics and hyperparameters used in our TFBS experiments are summarized in Table \ref{Tab:02}.

\begin{figure}[t]
\centering
\includegraphics[width=\columnwidth]{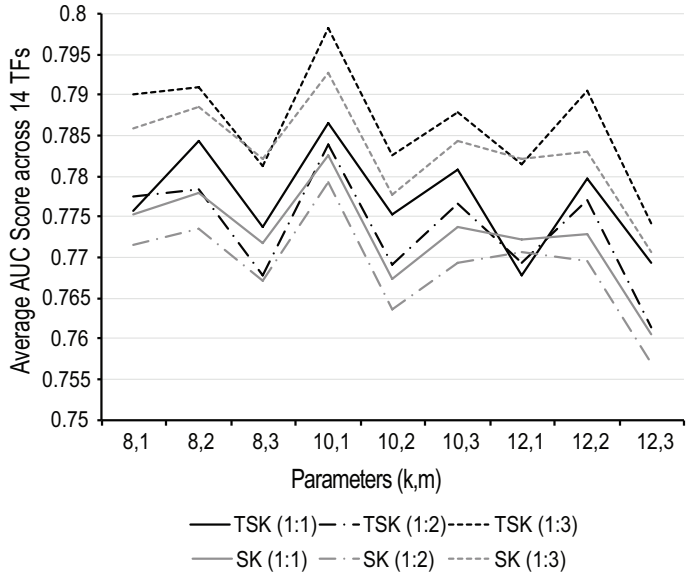}
\caption{Average AUC scores from TSK and SK on TFBS predictions when varying hyperparameters $(k,m)$. For each of the three data ratios (1:1,1:2 and 1:3) about validation sets, a curve is shown to describe the change in performance for SK and TSK with varying $(k,m)$ values.
Parameter combination of $k$=10 and $m=1$, gives the best performance for both SK and TSK. For each data ratio, TSK outperforms the basic SK.}
\label{fig:parameterTF}
\end{figure}

\begin{figure}[t]
\centering
\includegraphics[width=0.7\columnwidth]{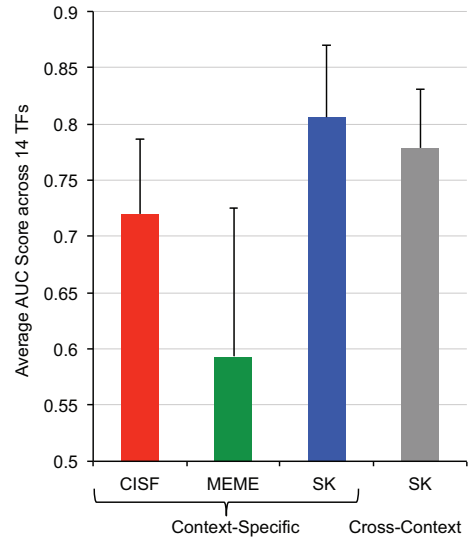}
\caption{Comparison using average test AUC scores of 14 TF tasks by comparing TFBS predictions under the Within-Context (``human Context-Specific'') setting versus Cross-Context setting. As expected, using the same SK approach, within-context setting achieves better performance than SK under cross-context setting indicating that the Cross-Context task is more difficult. Interestingly, our baseline SK method outperforms the state-of-art tools MEME and CISFINDER under the (easier) Within-Context (``human Context-Specific'') setting.}
\label{fig:controlTF}
\end{figure}

\subsection{Evaluation Metric} 
Our investigation of TSK focuses on the SVM \cite{joachims1999making} \cite{scholkopf1999advances} classifier framework. 
SVMs are known to provide state-of-the-art performance for
many applications \cite{VAPNIKLEARNING}, including a large
number of successes in
computational biology \cite{scholkopf2004kernel}. Binary SVM learns a real-valued 
function that assigns a continuous score to each candidate 
sequence segment based on the labeled source set. 
Larger scores correspond to more confident
predictions as the positive class. As a result, we rank all candidate samples
 in the target set using SVM prediction values.
This naturally leads us to utilize Area Under Curve (AUC) score (from the Receiver Operating Characteristic (ROC) curve) as our main evaluation metric.
The area under the ROC
curve (AUC) is commonly used as a summary measure of diagnostic
accuracy. It is interpreted as the probability that a randomly selected ``event'' will
be regarded with greater suspicion (in terms of its continuous
measurement) than a randomly selected ``non-event''. 
AUC score ranges between 0 and 1, where values 
closer to 1 indicate more successful predictions.

\subsection{Baseline approaches used for comparison}
For  TFBS prediction tasks, we compare TSK to the 
following baselines:
\begin{enumerate}
\item \textbf{SK}: represents the $(k,m)$-mismatch kernel implementation adapted from \cite{kuksa2008scalable}. We also ran spectrum kernel and found that its performed worse than $(k,m)$-mismatch SK (See Section 4.1)
\item \textbf{MEME}: is the state-of-the-art TFBS prediction tool that scores each sequence for potential TF binding events based on Position Weight Matrices (PWMs). The PWMs are obtained from source sequences using motif discovery algorithm MEME-ChIP\cite{machanick2011meme} and then these motifs are used for scoring individual target sequences using AMA tool \cite{buske2010assigning}. These tools are part of the MEME-Suite \cite{bailey2009meme}. 
\item \textbf{CISF}: CISFINDER \cite{sharov2009exhaustive} is another state-of-the-art toolbox for TFBS detection that uses an exhaustive search for 
 DNA motifs and outputs a Position Frequency Matrix (PFM). It then scores each target sequence 
based on the generated PFM. 
\end{enumerate}
The source and target sequences for all above baselines are same as 
those used in our TSK approach.

\subsection{MHC PB Prediction Task} 
To demonstrate the generalization of our model, we implement TSK on another group of cross-context prediction tasks: predict PB to MHC proteins from mice to humans. 
The mouse and human datasets are provided by the 2012 Machine Learning Competition in Immunology \cite{mlcompimmune}. The target human datasets have peptide bindings (PBs) for 5 different MHC molecules: HLA-A0201, HLA-B0702, HLA-B4403, HLA-B5301 and HLA-5701; while for the source mouse domain we select H2-Kb PB data. This leads to
5 different PB prediction tasks. We present 
the details of these datasets in Table \ref{Tab:03}. Important data statistics and hyperparameters of this task are listed as follows:
\begin{itemize}
\item \textbf{Dictionary size (d)}: Peptides are composed of 2 or more amino acids. Thus, the dictionary 
 size is much larger, with $d$=20, since there exist 20 amino acids. 
\item \textbf{Number of sequences (n)}: While the number of sequences of source domain is 3184, the number of sequences in target domain vary drastically for each MHC task. 
The details are included in Table \ref{Tab:03}. The length of these peptide sequences varies from 8 to 11 amino acids.
\item \textbf{Ratio of positive to negative samples in target data (r)}: The datasets are from the 2012 competition \cite{mlcompimmune}. The ratio between positive sites (i.e. sites with naturally processed (eluted) peptide binding to MHC-I complex) 
and the negative sites is already skewed (details in table \ref{Tab:03}).
\item \textbf{String kernel hyperparameters ($k$,$m$)}: Hyperparameter tuning was performed by training on mouse PB datasets and using the training data of human tasks as the validation set. We vary the length of the k-mer, 
$k \in \{5,6,7\}$, since the full sequence lengths are much smaller compared to TFBS,  and the number of allowed mismatches, $m \in \{1,2,3\}$. 
Model evaluation is done on human test datasets provided by the 2012 Machine Learning Competition in Immunology \cite{mlcompimmune}.
\item \textbf{Baselines}: Since no runnable tools exist for both oligo kernels and GS kernels, we use ($k,m$)-mismatch kernels from \cite{kuksa2008scalable} as a baseline.
\end{itemize}

\begin{figure}[t]
\centering
\includegraphics[width=\columnwidth]{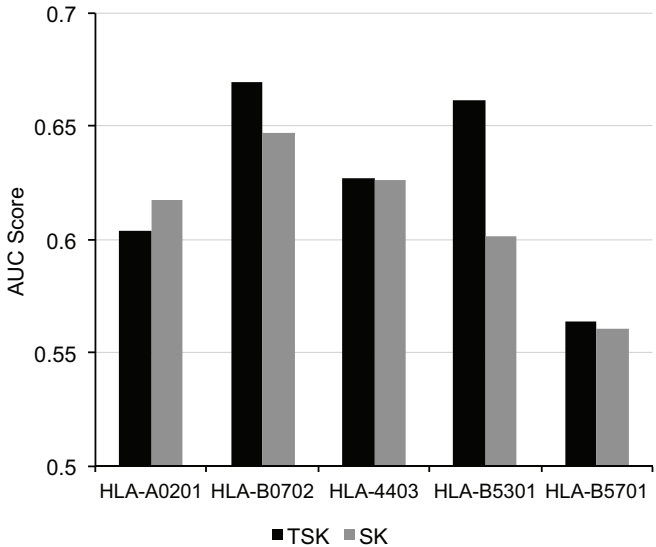}
\caption{Comparison using AUC scores across 5 tasks of MHC peptide binding (PB) prediction. We use mouse dataset as source and human dataset as target. TSK outperforms SK for 4 out of 5 cases.}
\label{fig:perfPB}
\end{figure}

\section{Results}

\subsection{Choice of Basic Kernel: $(k,m)$-mismatch kernel}
We select $(k,m)$-mismatch kernel as it allows mismatches during $k$-mer matching of biological sequences. Such sequences are
prone to mutations like substitutions, insertions, or deletions. In Table \ref{Tab:04a} we present average AUC scores from 14 TF tasks
for both $(k,m)$-mismatch kernel and spectrum kernel. As expected, the inexact matching of
 $(k,m)$-mismatch kernel is more effective and outperforms than the exact matching of spectrum kernel, making it a natural choice for our TSK approach.

\subsection{Evaluation of performance: TSK successfully transfers knowledge across different species}
Table \ref{Tab:04} presents the test AUC scores from different approaches for cross-context TFBS predictions of 14 selected TFs. For each TF, training was done on TFBS data 
from the mouse genome (source) and testing data was obtained from the human genome (target). 
We select $k=10$ and $m=1$ for SK and TSK based on validation set performance (details in Section~\ref{sec:hyperparam}).
TSK outperforms SK and two PWM baselines for most of the cases and is generally robust for imbalanced datasets (when r=1:2 and 1:3). 
Position Weight/Frequency Matrix based approaches, MEME and CISFINDER, 
also perform well for three TFs (USF2, RAD21 and SMC3). However, their performance is worse than SK and TSK for majority of the TFs. MEME and CISFINDER also 
exhibit a higher variance when considering all 14 TF prediction tasks.

Furthermore, we calculate the conservation scores for each of the 14 different TFs we used.
The conservation scores indicate how well conserved (or slowly evolving) individual positions of TFBS sequences are, for a particular TF. 
Conservation scores are calculated for 100 basepair-length sequences. Segments obtained from top 2000 binding sites of all 14 TFs
from the human genome. For these sequences, a phyloP score for each nucleotide (total of 200,000) is calculated using the Galaxy tool \cite{goecks2010galaxy,blankenberg2010galaxy,giardine2005galaxy}.
A positive phyloP score denotes conservation (slow evolution) and a negative phyloP score denotes acceleration (fast evolution) of the nucleotide.
Thus, after adding and normalizing the scores according to positively and negatively scored nucleotides for each TF, we take their log (for scaling) 
and subtract the acceleration scores from the conservation scores. We then include information regarding fraction of non-conserved nucleotides for each TF
to our final score. Non-conserved nucleotides are those nucleotides for which phyloP scores (positive or negative) are not reported. We calculate
this as a penalty term : 

\begin{equation}
p=\frac{\log(\frac{C_n}{C_t})}{100}
\end{equation}

The equation for calculating the final conservation score (CS) is:
\begin{equation}
CS = \log(PosScore) - \log(|NegScore|) - p
\end{equation}
where $C_n$ is the count of non-conserved nucleotides and $C_t$ is 
the count of total number of nucleotides $(=200,000)$.
$PosScore$ denotes normalized sum of all positive phyloP scores (quantifying slow evolution) while
$NegScore$ denotes normalized sum of all negative phyloP scores (quantifying fast evolution). Conservation scores for each TF are provided in the last column of Table \ref{Tab:04}.

For TFs that MEME and CISFINDER give high accuracy,
we observe that they also have high conservation scores. When the conservation scores are not high, SK and TSK models perform better.
Thus, MEME and CISFINDER perform well on TFs with strong sequence conservation across genomes.
This indicates that string kernel based approaches can be used to complement Position Weight/Frequency Matrix based approaches. 

It is important to note that TSK consistently outperforms SK even on more conserved TFs. This demonstrates that domain adaptation technique helps to improve the basic string
kernel for cross-context sequence predictions.

\subsection{Hyperparameter Selection: $k=10$ and $m=1$ gives the best performance for TFBS prediction}
\label{sec:hyperparam}
$(k,m)$-mismatch string kernel requires the tuning of two hyper parameters: $k$ and $m$. Here, $k$
is the length of the k-mer or substrings being compared in kernel calculations and $m$ is the number of mismatches being allowed.
We considered $k \in \{8,10,12\}$ and $m \in \{1,2,3\}$. 
Figure \ref{fig:parameterTF} shows the effect of varying $k$ and $m$ parameters 
using average AUC scores (across 14 TF tasks) from validation for all three considered data ratios. 
The combination ($k$=10, $m$=1) achieves the best average AUC results on validation sets.
We observe that for each particular ratio, TSK always outperforms basic SK on validation.
We also find that value $k$>10 decreases 
the performance of SK and TSK. 

\begin{figure*}[t]
 \centerline{\includegraphics[width=\textwidth]{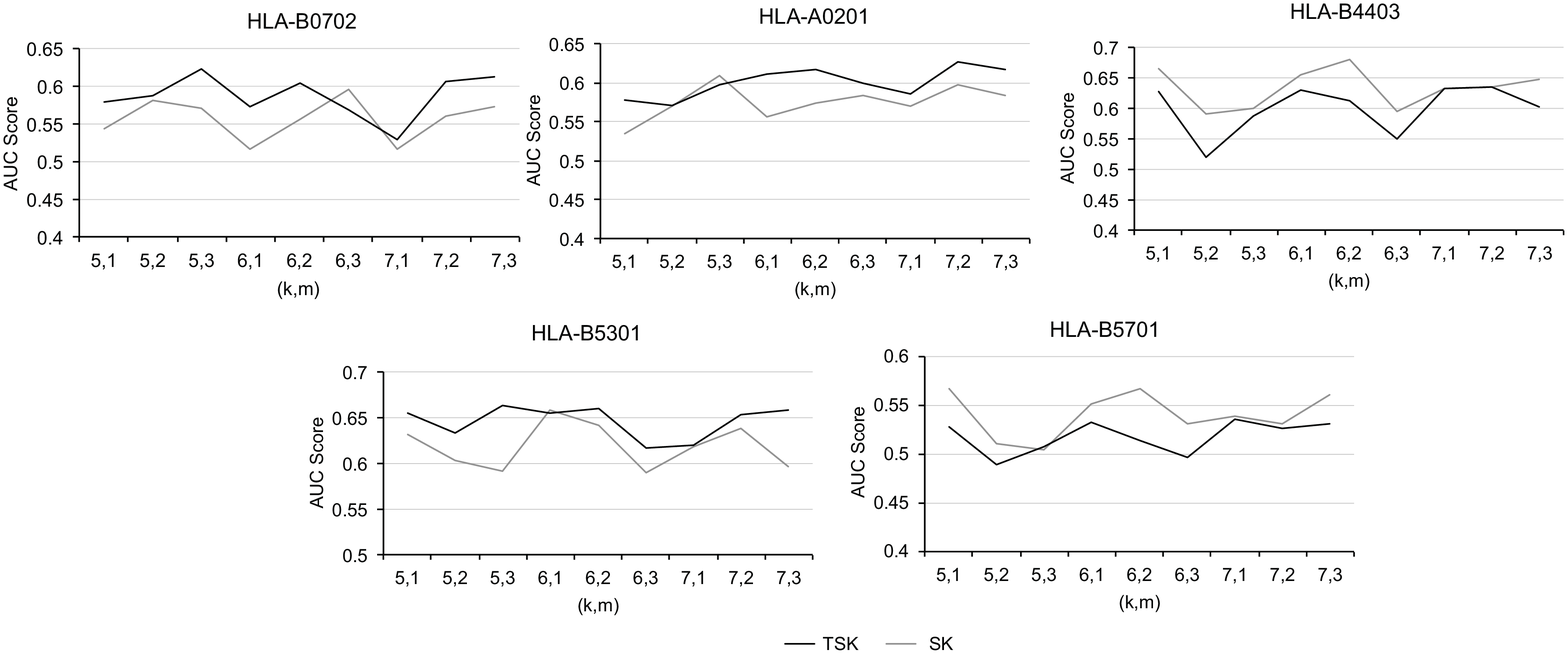}}
\caption{Average AUC scores on validation sets from TSK and SK when varying hyperparameters $(k,m)$ on all 5 MHC PB tasks.
Different best performing $k$ and $m$ values are selected across TSK and SK and across each task respectively.}
\label{fig:parameterPB}
\end{figure*}

\subsection{Same context versus cross-context: Prediction of TFBS within the target context is easier than cross-context}
To justify the importance of domain adaptation, we also evaluate baseline methods under the setting that train and test are within the same target domain. That is, both
the training and testing are performed on human TFBS datasets \footnote{We generate a new training set from human TFBS data (r=1:1)}. In Figure \ref{fig:controlTF},
the prediction performance, measured by average AUC test score, across 14 TF tasks, of SK decreases when under cross-context setting (in comparison
to the prediction from within the same context). However, it is worthwhile to point out that basic SK outperforms MEME and CISF for TFBS when working within the same domain. 

\begin{figure}[t]
\centering
\includegraphics[width=\columnwidth]{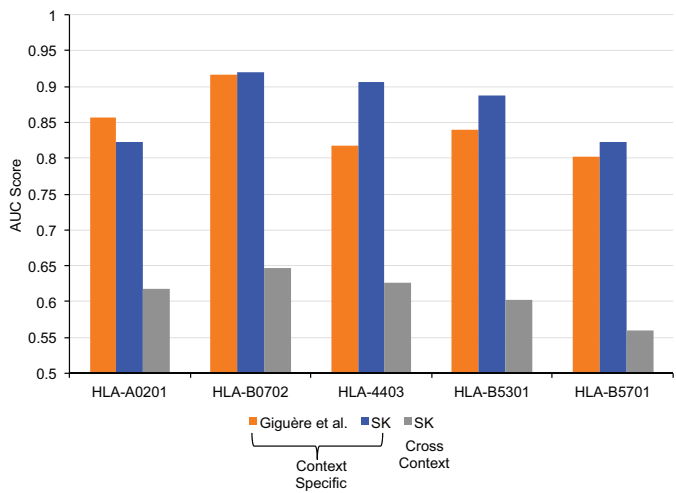}
\caption{Test AUC scores for the 5 MHC PB prediction tasks when comparing the Within-Context (human ``Context-Specific'') setting versus Cross-Context setting. As expected, SK approach, under the within-context setting, achieves better performance than SK under cross-context, indicating that Cross-Context tasks are more difficult. Moreover, our mismatch SK baseline outperforms Generic String Kernel ~\cite{giguere2013mhc} on 4 out of 5 tasks, making it a good choice for implementation of TSK on this type of task.}
\label{fig:controlPB}
\end{figure}

\subsection{TSK is generalizable to other sequence-based prediction tasks}
In addition to TFBS prediction, we successfully implement TSK for predicting PB to MHC-I complex by transferring knowledge across species. Similar to TFBS prediction, we use the mouse (source) dataset 
to train our model and then perform predictions (validation and testing) on the human (target) dataset, i.e. a useful translational setting. Our results indicate that TSK can be generalized to any cross-context task that involves sequence-based classification.
\begin{itemize}
\item \textbf{Evaluation of performance}: 
As shown in Figure \ref{fig:perfPB}, TSK performs better than the basic SK model 
on 4 out of 5 PB-prediction tasks across species.
\item \textbf{Hyperparameter Selection}: 
The hyperparamter tuning gave best-performing hyperparameters that were different across 5 tasks as well as across
TSK and SK approaches. Figure \ref{fig:parameterPB} presents tuning results for all 5 MHC PB prediction tasks, where training was done on the mouse H2-Kb
MHC PB dataset and validation was done on train data for human provided by 2012 competition in Immunology \cite{mlcompimmune}. 
Each task uses a different test dataset for human model evaluation.
The best performing  $k$ and $m$ values are selected for each task respectively.
\item \textbf{Same context versus cross-context}: Once again, we observe in Figure \ref{fig:controlPB} that the task of PB prediction is easier when done within the target (human) domain. That is, when both training and test datasets are from human domain. In this setting, our basic SK model (blue bars) gives good performance and even outperforms GS kernel ~\cite{giguere2013mhc}, the best team for 2012 competition cite{mlcompimmune}. This shows that SK is a good choice for implementation of domain adaptation technique. The grey bar in Figure \ref{fig:controlPB} shows that the performance of basic SK decreased drastically when predictions are made across contexts. Therefore, indicating the need to propose TSK for cross-context settings. \footnote{We also apply TSK to this within-context setting and get following AUC scores: 0.8325 (HLA-A0201), 0.9233 (HLA-B0702), 0.9282 (HLA-B4403), 0.8599 (HLA-B5301) and 0.82 (HLA-B5701). Interestingly, we find that TSK adapts and reduces the differences among the training and testing datasets and gives slightly better performance on 3 out of 5 MHC PB prediction tasks (HLA-A0201, HLA-B0702 and HLA-B4403). SK's performance is represented by blue bars in Figure \ref{fig:controlPB}}
\end{itemize}

\section{Discussion}

Determining how TF proteins interact with DNA to regulate context specific gene regulation is essential for fully understanding biological processes and diseases.  Most previous computational tools for predicting TFBSs assume the same distribution across target and source contexts. We, however, have shown that it is beneficial to consider the distribution shift. The proposed TSK method improves the performance of sequence-driven TFBS predictions by accommodating differences 
among underlying sample distributions and applying knowledge transfer from the source to target context.  
We have also examined the imbalanced (positive to negative ratio) data issue in the target domain, 
to ensure realistic and robust TFBS predictions. 
Our experimental results indicate:
\begin{itemize}
\item TSK overall improves performance of string kernel on cross-context TFBS predictions;
\item TSK and SK outperform the state-of-the-art Position Weight/Frequency Matrix based 
TFBS tools, especially for less conserved TFs, and can be used as complementary tools to the latter;
\item TSK can be easily generalized to other sequence based prediction tasks, e.g. prediction of PB to MHC-I complex molecules.
\end{itemize}

The code for TSK has been made available at \url{www.github.com/QData/TransferStringKernel}.
TSK is a general sequence modeling architecture.
Therefore, our future plan is to extend TSK to a wider variety of sequence based applications. 
We would also like to study and compare implementation of other kernels from 
the string kernel family as components of the TSK approach.
In the meantime, we are aware of the limitations of TSK due to its high computational complexity 
when handling datasets with a large number of source and target samples. We plan to explore  
random sampling based techniques to scale up TSK on ``large-scale" sequence mining tasks. 

\ifCLASSOPTIONcompsoc

  \section*{Acknowledgments}
\else

  \section*{Acknowledgment}
\fi
The authors thank Dr. Mazhar Adli (Department of Biochemistry and Molecular Genetics, University of Virginia) for helpful discussions throughout the progress of this project.

\ifCLASSOPTIONcaptionsoff
  \newpage
\fi

\section*{Funding}
This work was supported by the National Science Foundation under NSF
CAREER award No. 1453580. Any Opinions, findings and conclusions or recommendations expressed
in this material are those of the author(s) and do not necessarily reflect those of the National Science
Foundation.

\bibliographystyle{IEEEtran}
\bibliography{TSK-BIOKDD}

\newpage
\begin{IEEEbiographynophoto}{Ritambhara Singh}
is a $4^{th}$ year PhD student at University of Virginia. She completed her Bachelors in Computer Engineering from University of Pune, India. She joined the Computer Science Department in July 2012, and has been working on applying machine learning algorithms to novel biological applications. In order to understand the biological data and related tasks, she worked closely with members of Adli Lab in the Department of Biochemistry and Molecular Genetics from 2012 to 2015. Her current research involves developing efficient algorithms for bio-sequence analysis.
\end{IEEEbiographynophoto}
\vspace{-1cm}
\begin{IEEEbiographynophoto}{Jack Lanchantin}
is a $2^{nd}$ year PhD student in the Computer Science Department at University of Virginia. He completed his Bachelors in Computer Engineering from the State University of New York at Binghamton. His research focuses on applying machine learning for biomedical applications. 
\end{IEEEbiographynophoto}
\vspace{-1cm}
\begin{IEEEbiographynophoto}{Gabriel Robins}
is a Professor in the Department of Computer Science, at the University of Virginia, where he received an NSF Young Investigator Award, a Packard Foundation Fellowship, the SIAM Outstanding Paper Prize, a University Teaching Fellowship, an All-University Outstanding Teaching Award, a Faculty Mentor Award, a two-year early promotion/tenure, the Walter N. Munster Endowed Chair, and the Virginia Engineering Foundation Faculty Appreciation Award. He completed his Ph.D. in Computer Science in 1992 at UCLA, where he received an IBM Fellowship and a Distinguished Teaching Award. Gabe's primary area of research is VLSI CAD, and he co-authored a book on high-performance routing. His additional research interests include algorithms, RFID, bioinformatics, computational geometry, combinatorial optimization, and computational biology. 
\end{IEEEbiographynophoto}
\vspace{-1cm}
\begin{IEEEbiographynophoto}{Yanjun Qi}
Dr.Yanjun Qi is an assistant professor of University of Virginia, Department of Computer Science since 2013. She was a researcher in the Machine Learning Department at NEC Labs American, Princeton, NJ from July 2008 to August 2013. Her research interests are within machine learning, data mining, and bioinformatics. She obtained her Ph.D. degree from School of Computer Science at Carnegie Mellon University in May 2008 and received her Bachelor degree with high honors from Computer Science Department at Tsinghua University, Beijing. She has served as PCs and reviewers for multiple renowned international conferences/ journals, and has co-chaired the NIPS ``Machine Learning for Computational Biology'' workshops. Dr. Qi has received CAREER award from NSF and a Best Paper Award at International Conference of BodyNet.
\end{IEEEbiographynophoto}

\enlargethispage{-5in}

\end{document}